# Motor Imagery Classification based on CNN-GRU Network with Spatio-Temporal Feature Representation[*]

Ji-Seon Bang[1] and Seong-Whan Lee[2]

[1] Department of Brain and Cognitive Engineering
[2] Department of Artificial Intelligence
Korea University, Seoul, Republic of Korea
{js_bang,sw.lee}@korea.ac.kr

**Abstract.** Recently, various deep neural networks have been applied to classify electroencephalogram (EEG) signal. EEG is a brain signal that can be acquired in a non-invasive way and has a high temporal resolution. It can be used to decode the intention of users. As the EEG signal has a high dimension of feature space, appropriate feature extraction methods are needed to improve classification performance. In this study, we obtained spatio-temporal feature representation and classified them with the combined convolutional neural networks (CNN)-gated recurrent unit (GRU) model. To this end, we obtained covariance matrices in each different temporal band and then concatenated them on the temporal axis to obtain a final spatio-temporal feature representation. In the classification model, CNN is responsible for spatial feature extraction and GRU is responsible for temporal feature extraction. Classification performance was improved by distinguishing spatial data processing and temporal data processing. The average accuracy of the proposed model was 77.70% (± 15.39) for the BCI competition IV_2a data set. The proposed method outperformed all other methods compared as a baseline method.

**Keywords:** Brain-computer interface (BCI) · Electroencephalography (EEG) · Motor imagery (MI) · Convolutional neural network (CNN) · Gated recurrent unit (GRU).

## 1 Introduction

Brain-computer interfaces (BCI) allows users to control external devices with their intentions, which are decoded from users' brain signals [1–5]. Motor im-agery (MI) tasks are widely used for BCI paradigms. Sensory motor rhythms [6–8] are induced when humans mentally simulate certain movements in their minds. Whenever a subject imagines the movement of the particular body part, the corresponding parts of the brain are activated. This brain activity can be recorded and used to control external devices.

[*] This work was partly supported by Institute of Information & Communications Technology Planning & Evaluation (IITP) grant funded by the Korea government (MSIT) (No. 2015-0-00185, Development of Intelligent Pattern Recognition Softwares for Ambulatory Brain Computer Interface, No. 2017-0-00451, Development of BCI based Brain and Cognitive Computing Technology for Recognizing User's Intentions using Deep Learning, No. 2019-0-00079, Artificial Intelligence Graduate School Program(Korea University)).

There are many studies to achieve better classification performance of MI. Common spatial pattern (CSP) [9] and its variants are the most commonly used feature extraction methods. CSP finds the set of spatial filters that maximizes the distance of variance for multiple classes [10]. Common spatio-spectral pattern (CSSP) [11] is extended versions of CSP. It provides a temporal delay to obtain various features.

In the process of decoding the electroencephalography (EEG) signal for the BCI system, feature extraction is particularly important. As MI signals have high variability between and within subjects, decoding accuracy can be improved by extracting the subject-specific features of each subject [12–14]. However, performance can often be reduced because this part is usually not well considered [15].

In the process of decoding the electroencephalography (EEG) signal for the BCI system, feature extraction is particularly important. As MI has large inter- and intra-subject variability, classification performance can be potentially enhanced by extracting features with subject-specific parameters [12, 13]. However, they are often selected heuristically [15], which can cause poor classification performance.

Recently, the use of neural network methods has increased in the BCI field [16–22], for various BCI classification tasks. Cecotti *et al.* [16], Manor *et al.* [17], and Sturm *et al.* [18] focused on the EEG time series of each channel. Stober *et al.* [19] and Bashivan *et al.* [20], targeted frequency components using fast Fourier transform (FFT) and short-time Fourier transform (STFT). Sakhavi *et al.* [21] adopted spatial filter on the EEG time series data and Bang *et al.* [22] focused on spatio-spectral domain of the EEG signal.

Here, CNN is one of the artificial neural network models that has several convolutional layers and a fully connected layer. The neurons in the CNN jointly implement a complex nonlinear mapping by adapting the weights of each neuron. The CNN was first designed to recognize two-dimensional images [23], and the network functions by convolving the images with a two-dimensional kernel in the convolutional layer.

On the other hand, unlike CNNs specialized in dealing with spatial information, there also exist RNN, which is specialized in dealing with temporal information. Among them, GRU is a type of RNN, the network of RNN is connected along the temporal sequence. As the connections are formed between nodes like a directed graph, it allows handling temporal dynamics. The difference between other types of RNN is that GRU includes forgetting gate with fewer parameters.

Here, we propose a method for classifying MI signals with combined CNN and gated recurrent unit (GRU) network. We adopted CNN to handle spatial information and GRU to handle temporal information. To generate input features for the classification, we first adopted spatial feature representation with normalized sample covariance matrix (NSCM) [24] for each temporal band. Then we concatenated them to generate the final input feature map for the CNN-GRU network.

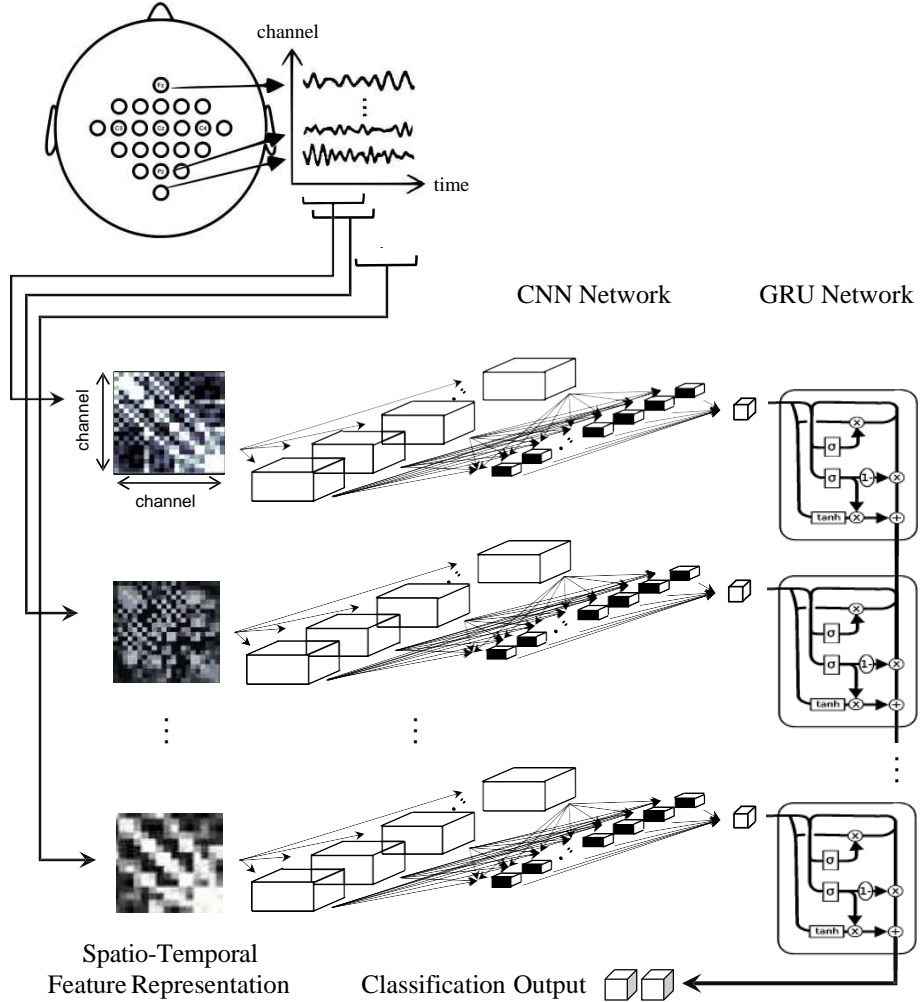

**Fig. 1.** The framework of the proposed CNN-GRU network with spatio-temporal feature representation.

The concept of the proposed method is similar to CSSP, which provides a temporal delay to obtain various features. However, unlike CSP, our method extracts spatial and temporal information separately. First, the CNN network extract features from NSCM feature maps within each time point. The reason for adopting CNN is that the NSCM feature maps contain spatial information. From this process, a key feature value is extracted at each point. After that, the GRU network extract features on the time domain. As the GRU network is specialized for classification on time-series, we adopted it for the second process. By dividing the steps of feature extraction, we were able to further refine classification to improve performance.

## 2  Method

In this section, we will introduce what we considered for the proposed method. Also, we will give a brief explanation of the EEG signal. Specifically, we will first focus on how to generate an input matrix that contains spatial and temporal information from each of the EEG trials. Then we will explain the design of the CNN-GRU network that can handle spatio-temporal feature representation. Fig. 1 presents the overview of the proposed method. Further detailed information will be explained below.

### 2.1  Spatio-Temporal Feature Generation

When analyzing EEG data, bandpass filtering is a common pre-processing method. It removes unnecessary artifacts and specifies the signal of interest. for the filtering band, we specified a common frequency range of 8-30 Hz, so that it can be applied equally to all subjects. 8-30 Hz includes alpha-band and beta-band, which are brain signals that are active when a person is awake and focused. It is also a commonly used frequency band when classifying MI signals.

The raw MI signal is filtered by 8-30 Hz. After that, we segmented the filtered signal between 0.5 sec and 2.5 sec to 2 sec and 4 sec after the onset of the cue with 0.1 sec sliding window. With this process, both spatial information and temporal information can be preserved.

The segmented signal is normalized by adopting a function named local average reference. The reference channel for the function was electrode Cz. Then, the normalized signal was compressed with NSCM. NSCM is a type of covariance matrix. In general, the covariance matrix is known to extract the informative features well in MI classification. For this reason, we generated input feature representation for the CNN-GRU network in covariance form. The NSCM matrices which are generated for each temporal band are concatenated. This spatio-temporal feature representation is finally adopted as the input for the CNN-GRU network. The size of feature representation is $C \times C \times T$, as the size of NSCM is $C \times C$. $C$ denotes the number of channels and $T$ denotes the size of the temporal band.

### 2.2  Classification with CNN-GRU Network

To decode spatio-temporal feature representation, we designed a CNN-GRU network. In the network, two 2-dimensional convolution layers were adopted to handle spatial features, and GRU was utilized further to process temporal information which is sequential feature map output from convolution layers. To be specific, two convolutional layers learn $C \times C$ sized spatial features and one GRU layer learns $T$ time step temporal features. The network configuration is summarized in Table 1.

For the network, we applied two convolutional layers of size ($K \times K$) and (($C - K + 1) \times (C - K + 1)$), respectively, to learn spatial information from the input. $K$ is the kernel size, which is set to 3. For all convolutional layers, we adopted bias and rectified linear units (RELU) [25]. The outputs from these two convolutional layers of temporal step size $T$ are put to the GRU layer to learn temporal information. Between last convolutional layer and GRU layer, dropout [26] layers are adopted with a probability of 80%. The final classification is performed from the last GRU layer. We fit the model using Adam-optimizer [27] to optimize the cost. For training, the batch size was set equal to the test set size, which is called the full-batch setting. The performance was obtained when the number of epochs reached 500, and the learning rate was 0.0001.

**Table 1.** Configuration of CNN-GRU Network.

| Layer | #1 Conv layer | #2 Conv layer | #3 Dropout | #4 GRU |
|---|---|---|---|---|
| Input size | $C \times C \times T$ | $(C-K+1) \times (C-K+1) \times T$ | $1 \times 1 \times T$ | $1 \times 1 \times T$ |
| Kernel size | $K \times K$ | $(C-K+1) \times (C-K+1)$ | - | - |
| Padding | 0 | 0 | - | - |
| Stride | 1 | 1 | - | - |
| Outputs | 128 | $T$ | $T$ | #class labels |

## 3  Data Description and Evaluation

### 3.1  BCI Competition IV_2a Data

We verified the proposed method with BCI Competition IV_2a data [28], which is commonly used for verification of EEG motor imagery studies. The data set consists of the EEG data collected from 9 subjects, namely (A01–A09). The data consist of 25 channels, which include 22 EEG channels, and 3 EOG channels with a sampling frequency of 250 Hz. EOG was not considered in this study. The data were collected on four different motor imagery tasks but only left and right-hand motor imagery tasks are selected for this study. The EEG data were sampled at 250 Hz and band-pass filtered between 0.5 Hz and 100 Hz. To suppress line noise, a 50 Hz notch filter was also adopted. The performance of the proposed method was evaluated by accuracy. We followed the common practice in machine learning to partition the data into training and test sets. The total number of 144 trials for two tasks were used as the training and the same number of trials are used for the test set. In this study, the experiments were conducted in the Tensorflow environment on Intel 3.20 GHz Core i5 PC with 24 GB of RAM.

### 3.2   Baseline Methods

We compared the classification accuracy of our model with baseline methods. CSP [9] and CSSP [11] were employed as the linear methods for the feature extraction method. For the classification, LDA was applied to all linear methods [29, 30]. The signal was band-pass filtered between 8 and 13 Hz ($\mu$ band).

We also compared our proposed method with the nonlinear methods described by Sturm *et al.* [18] and Sakhavi *et al.* [21]. For comparison, the same pre-processing methods used in the corresponding papers were adopted. To compare with Sturm *et al.* [18], the signal was band-pass filtered between 9 and 13 Hz. The signal for comparisons with Sakhavi *et al.* [21] was band-pass filtered with a filter bank containing nine subsequential filters (4–8 Hz, 8–12 Hz, ...), with four of them being selected. The first to the fourth were selected, after the subsequential filters are sorted in descending order of the mutual information, as per the corresponding paper. The signal that was segmented between 0.5 and 2.5 s after the onset of the cue was used for segmentation for both methods [31, 32]. After segmentation, we adopted a local-average-reference function on the Sturm *et al.* [18]. As the feature extraction methods used by Sakhavi *et al.* [21] are based on CSP, these functions were not adopted in this method. For both methods, the learning rate was set to 0.0001, full-batch setting was adopted, and the initial value of the weight was fixed. The final accuracy was determined when the number of iterations of the classifier reached 500, as in the proposed method. Note that in the method used by Sakhavi *et al.* [21], parameters such as kernel size and the number of hidden nodes are optimized. Hence, the same parameters as in the corresponding paper were used for comparison.

## 4   Results

We compared the proposed model with baseline methods by obtaining decoding accuracy. Performance was obtained from given training data and test data of the BCI competition IV_2a data.

### 4.1   Comparison with Baseline Methods

Table 2 represents the performance that were obtained with the BCI Competition IV_2a data. Results of both proposed method and previous linear and nonlinear methods are presented for all subjects. The mean accuracy of the baseline methods were 72.84% (±16.93), 73.38% (±19.88), 73.53% (±15.70), 76.60% (±6.74) for CSP [9], CSSP [11], Sturm *et al.* [18], and Sakhavi *et al.* [33] respectively. The mean accuracy of the proposed model was 77.70% (±15.39) across subjects. The proposed method outperformed all baselines methods.

**Table 2.** Comparison of the proposed method and baseline method with BCI Competition IV_2a Data.

| Methods<br>Subjects | CSP<br>[9] | CSSP<br>[11] | Sturm<br>*et al.* [18] | Sakhavi<br>*et al.* [21] | Proposed |
|---|---|---|---|---|---|
| A01 | 79.86 | 88.89 | 72.92 | 75.00 | 88.8 |
| A02 | 49.31 | 51.39 | 63.19 | 63.19 | 54.86 |
| A03 | 97.22 | 94.44 | 94.44 | 78.47 | 97.22 |
| A04 | 59.03 | 52.08 | 60.42 | 77.78 | 72.22 |
| A05 | 59.72 | 50.69 | 56.94 | 82.64 | 61.11 |
| A06 | 66.67 | 61.81 | 65.97 | 70.83 | 67.36 |
| A07 | 63.89 | 72.22 | 61.81 | 80.56 | 73.61 |
| A08 | 93.75 | 95.83 | 96.53 | 86.11 | 95.83 |
| A09 | 86.11 | 93.06 | 89.58 | 75.69 | 88.19 |
| mean | 72.84 | 73.38 | 73.53 | 76.60 | 77.70 |

### 4.2  Comparison Spatial Feature Representation Only

We investigated whether our proposed method is appropriate. If the performance is enhanced when using only spatial feature representation, there will be no need for the spatio-temporal feature representation. In our proposed method, spatial features maps are extracted through NSCM from the continuous temporal band and they are concatenated. Total of 16 temporal bands were used at 0.1 second intervals from 2.5-4.5 to 4.0-6.0. For comparison, each of these 16 NSCM feature representations was classified using only CNN and then compared to the proposed method with decoding accuracy. In this process, only the GRU layer was excluded.

Table 3 and Fig. 2 shows the results. Table 3 shows detailed decoding accuracy of each temporal band and each subject. In addition, Fig. 2 graphically represents only the average performance of the nine subjects by each temporal band for legibility. In the figure, bars represent the average performance of nine subjects in each temporal band, and red lines represent the performance of the proposed method. Seeing the result, the minimum accuracy was 68.06% and the maximum accuracy was 77.55%. In conclusion, no performance was higher than the 77.70% which is the performance of the proposed method throughout the entire 16 temporal bands.

**Table 3.** Decoding accuracy when using spatial feature representations only

| Sub.<br>Temp. | A01 | A02 | A03 | A04 | A05 | A06 | A07 | A08 | A09 | mean |
| --- | --- | --- | --- | --- | --- | --- | --- | --- | --- | --- |
| 2.5-4.5 | 89.58 | 53.47 | 96.53 | 66.67 | 65.97 | 70.83 | 72.92 | 96.53 | 84.03 | 77.39 |
| 2.6-4.6 | 89.58 | 57.64 | 95.83 | 69.44 | 68.06 | 65.28 | 65.97 | 97.22 | 84.72 | 77.08 |
| 2.7-4.7 | 86.81 | 54.17 | 97.22 | 70.14 | 68.75 | 67.36 | 70.14 | 96.53 | 86.81 | 77.55 |
| 2.8-4.8 | 86.81 | 58.33 | 95.83 | 66.67 | 62.50 | 64.58 | 70.83 | 95.83 | 88.89 | 76.70 |
| 2.9-4.9 | 85.42 | 57.64 | 95.83 | 72.22 | 61.11 | 70.14 | 73.61 | 95.83 | 85.42 | 77.47 |
| 3.0-5.0 | 80.56 | 59.72 | 95.83 | 69.44 | 63.19 | 64.58 | 73.61 | 95.83 | 81.94 | 76.08 |
| 3.1-5.1 | 81.94 | 61.11 | 95.83 | 72.22 | 63.19 | 65.97 | 72.92 | 95.83 | 81.25 | 76.70 |
| 3.2-5.2 | 84.03 | 58.33 | 94.44 | 70.14 | 61.81 | 66.67 | 72.92 | 95.14 | 81.94 | 76.16 |
| 3.3-5.3 | 84.03 | 56.25 | 93.06 | 69.44 | 57.64 | 64.58 | 74.31 | 92.36 | 81.94 | 74.85 |
| 3.4-5.4 | 85.42 | 48.61 | 93.75 | 69.44 | 61.81 | 66.67 | 68.75 | 93.06 | 81.94 | 74.38 |
| 3.5-5.5 | 81.94 | 50.69 | 92.36 | 65.97 | 58.33 | 66.67 | 66.67 | 91.67 | 80.56 | 72.76 |
| 3.6-5.6 | 82.64 | 52.78 | 92.36 | 66.67 | 56.25 | 63.19 | 65.97 | 90.28 | 75.00 | 71.68 |
| 3.7-5.7 | 86.11 | 52.08 | 88.89 | 65.97 | 58.33 | 60.42 | 65.97 | 88.19 | 66.67 | 70.29 |
| 3.8-5.8 | 84.72 | 50.00 | 88.19 | 63.89 | 51.39 | 59.72 | 64.58 | 87.50 | 64.58 | 68.29 |
| 3.9-5.9 | 81.94 | 50.00 | 88.19 | 64.58 | 54.17 | 60.42 | 61.11 | 86.81 | 66.67 | 68.21 |
| 4.0-6.0 | 79.17 | 54.86 | 88.19 | 68.06 | 50.69 | 59.72 | 63.19 | 84.72 | 63.89 | 68.06 |

## 5  Discussion and Conclusion

In this study, a novel framework that includes spatio-temporal feature representation and corresponding CNN-GRU network was proposed for EEG signal classification. The proposed feature representation can retain structural dependencies of the task-related information of motor imagery based EEG signal. Here, we were able to demonstrate the superiority of the proposed method compared to other methods. Table 2 shows that the proposed method has higher decoding accuracy compared to other methodologies when compared with BCI Competition IV_2a data. We showed better performance than linear methods as well as nonlinear methods.

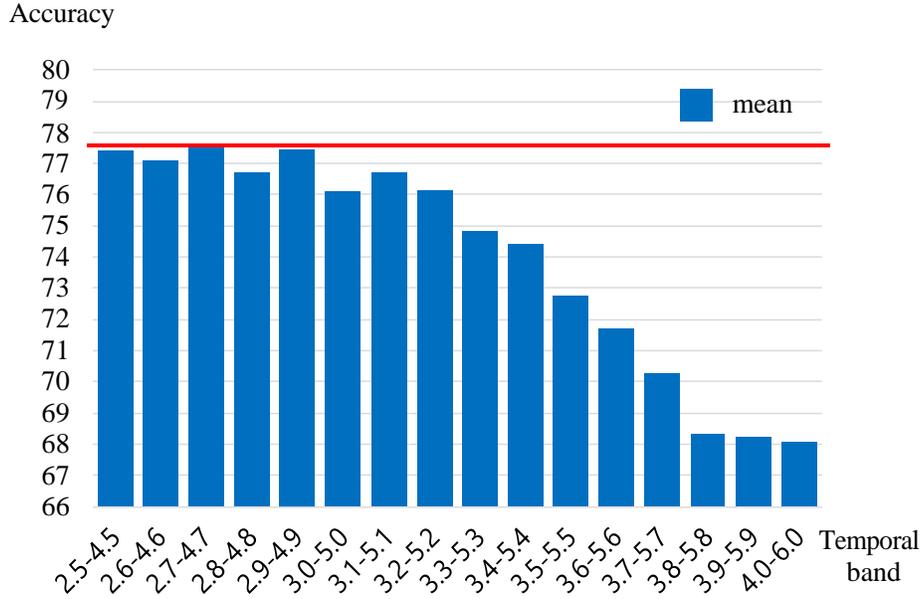

**Fig. 2.** Average decoding accuracy when using spatial feature representations only. The blue bars are average accuracy of each nine subjects with CNN and the red line indicates the performance of proposed CNN-GRU network.

We also tried to derive classification results only through spatial feature representation to determine whether the proposed method is meaningful. If the spatial feature representation with the CNN model achieves higher performance than the proposed method, it can be hard to say that the proposed method is useful. For this purpose, we obtained decoding accuracy via CNN without GRU, using only the spatial feature of each temporal band with NSCM. The result was that the performance for each temporal band was not as high as the proposed method at all. This shows that the proposed method that using spatio temporal information at once with the CNN-GRU network is better than simply using spatial feature information.

Generating feature representations by preserving spatial and temporal information of EEG signals has not been adequately addressed in previous CNN-based BCI studies. By preserving spatio-temporal relationships of EEG signals, proposed model has enabled to outperform previous approaches in spatial as

10     J.-S. Bang et al.well as temporal domains. To deal with temporal information, we adopted GRU networks as well as the CNN, achieving higher performance than using spatial information only. This demonstrated that the proposed method is useful. The proposed feature representation can be applied to other research areas where preserving spatio-temporal data dependencies of task-related information is important.

## References

1. Pfurtscheller, G., Neuper, C.: Motor imagery and direct brain-computer communication. Proceedings of the IEEE **89**(7), 1123–1134 (2001)
2. Chen, Y., Atnafu, A.D., Schlattner, I., Weldtsadik, W.T., Roh, M.C., Kim, H.J., Lee, S.W., Blankertz, B., Fazli, S.: A high-security EEG-based login system with rsvp stimuli and dry electrodes. IEEE Transactions on Information Forensics and Security **11**(12), 2635–2647 (2016)
3. Won, D.O., Hwang, H.J., Kim, D.M., Müller, K.R., Lee, S.W.: Motion-based rapid serial visual presentation for gaze-independent brain-computer interfaces. IEEE Transactions on Neural Systems and Rehabilitation Engineering **26**(2), 334–343 (2017)
4. Lee, M.H., Williamson, J., Won, D.O., Fazli, S., Lee, S.W.: A high performance spelling system based on eeg-eog signals with visual feedback. IEEE Transactions on Neural Systems and Rehabilitation Engineering **26**(7), 1443–1459 (2018)
5. Lee, M.H., Kwon, O.Y., Kim, Y.J., Kim, H.K., Lee, Y.E., Williamson, J., Fazli, S., Lee, S.W.: EEG dataset and OpenBMI toolbox for three BCI paradigms: an investigation into BCI illiteracy. GigaScience **8**(5), giz002 (2019)
6. Yuan, H., He, B.: Brain–computer interfaces using sensorimotor rhythms: Current state and future perspectives. IEEE Transactions on Biomedical Engineering **61**(5), 1425–1435 (2014)
7. Pfurtscheller, G., Brunner, C., Schlögl, A., da Silva, F.L.: Mu rhythm (de)synchronization and EEG single-trial classification of different motor imagery tasks. NeuroImage **31**(1), 153 – 159 (2006)
8. Suk, H.I., Lee, S.W.: Subject and class specific frequency bands selection for multiclass motor imagery classification. International Journal of Imaging Systems and Technology **21**(2), 123–130 (2011)
9. Ramoser, H., Muller-Gerking, J., Pfurtscheller, G.: Optimal spatial filtering of single trial EEG during imagined hand movement. IEEE Transactions on Rehabilitation Engineering **8**(4), 441–446 (2000)
10. Blankertz, B., Tomioka, R., Lemm, S., Kawanabe, M., Müller, K.R.: Optimizing spatial filters for robust EEG single-trial analysis. IEEE Signal Processing Magazine **25**(1), 41–56 (2008)
11. Lemm, S., Blankertz, B., Curio, G., Müller, K.R.: Spatio-spectral filters for improving the classification of single trial EEG. IEEE Transactions on Biomedical Engineering **52**(9), 1541–1548 (2005)
12. Krauledat, M., Tangermann, M., Blankertz, B., Müller, K.R.: Towards zero training for brain-computer interfacing. PloS One **3**(8), e2967 (2008)
13. Fazli, S., Popescu, F., Danóczy, M., Blankertz, B., Müller, K.R., Grozea, C.: Subject-independent mental state classification in single trials. Neural Networks **22**(9), 1305–1312 (2009)

14. Kwon, O.Y., Lee, M.H., Guan, C., Lee, S.W.: Subject-independent brain-computer interfaces based on deep convolutional neural networks. IEEE Transactions on Neural Networks and Learning Systems (2019)
15. Ang, K.K., Chin, Z.Y., Zhang, H., Guan, C.: Mutual information-based selection of optimal spatial–temporal patterns for single-trial EEG-based BCIs. Pattern Recognition **45**(6), 2137–2144 (2012)
16. Cecotti, H., Graser, A.: Convolutional neural networks for P300 detection with application to brain-computer interfaces. IEEE Transactions on Pattern Analysis and Machine Intelligence **33**(3), 433–445 (2011)
17. Manor, R., Geva, A.B.: Convolutional neural network for multi-category rapid serial visual presentation BCI. Frontiers in Computational Neuroscience **9**, 146 (2015)
18. Sturm, I., Lapuschkin, S., Samek, W., Müller, K.R.: Interpretable deep neural networks for single-trial EEG classification. Journal of Neuroscience Methods **274**, 141–145 (2016)
19. Stober, S., Cameron, D.J., Grahn, J.A.: Using convolutional neural networks to recognize rhythm stimuli from electroencephalography recordings. In: Advances in Neural Information Processing Systems. pp. 1449–1457 (2014)
20. Bashivan, P., Rish, I., Yeasin, M., Codella, N.: Learning representations from EEG with deep recurrent-convolutional neural networks. arXiv preprint arXiv:1511.06448 (2015)
21. Sakhavi, S., Guan, C., Yan, S.: Learning temporal information for brain-computer interface using convolutional neural networks. IEEE Transactions on Neural Networks and Learning Systems **29**(11), 5619–5629 (2018)
22. Bang, J.S., Lee, M.H., Fazli, S., Guan, C., Lee, S.W.: Spatio-spectral feature representation for motor imagery classification using convolutional neural networks. IEEE Trans. Neural Netw. Learn. Syst. (Jan 2021)
23. LeCun, Y., Boser, B.E., Denker, J.S., Henderson, D., Howard, R.E., Hubbard, W.E., Jackel, L.D.: Handwritten digit recognition with a back-propagation network. In: Advances in Neural Information Processing Systems. pp. 396–404 (1990)
24. Barachant, A., Bonnet, S., Congedo, M., Jutten, C.: Multiclass brain–computer interface classification by riemannian geometry. IEEE Transactions on Biomedical Engineering **59**(4), 920–928 (2011)
25. Nair, V., Hinton, G.E.: Rectified linear units improve restricted boltzmann machines. In: Proceedings of the 27th International Conference on Machine Learning (ICML-10). pp. 807–814 (2010)
26. Srivastava, N., Hinton, G., Krizhevsky, A., Sutskever, I., Salakhutdinov, R.: Dropout: a simple way to prevent neural networks from overfitting. The Journal of Machine Learning Research **15**(1), 1929–1958 (2014)
27. Kingma, D.P., Ba, J.: Adam: A method for stochastic optimization. arXiv preprint arXiv:1412.6980 (2014)
28. Tangermann, M., Müller, K.R., Aertsen, A., Birbaumer, N., Braun, C., Brunner, C., Leeb, R., Mehring, C., Miller, K.J., Mueller-Putz, G., et al.: Review of the BCI competition IV. Frontiers in Neuroscience **6**, 55 (2012)
29. Coyle, D., Satti, A., Prasad, G., McGinnity, T.M.: Neural time-series prediction preprocessing meets common spatial patterns in a brain-computer interface. In: Engineering in Medicine and Biology Society, 2008. EMBS 2008. 30th Annual International Conference of the IEEE. pp. 2626–2629. IEEE (2008)
30. Coyle, D.: Neural network based auto association and time-series prediction for biosignal processing in brain-computer interfaces. IEEE Computational Intelligence Magazine **4**(4) (2009)

31. Lotte, F., Guan, C.: Regularizing common spatial patterns to improve BCI designs: Unified theory and new algorithms. IEEE Transactions on Biomedical Engineering **58**(2), 355–362 (2011)
32. Ang, K.K., Chin, Z.Y., Wang, C., Guan, C., Zhang, H.: Filter bank common spatial pattern algorithm on BCI competition IV datasets 2a and 2b. Frontiers in Neuroscience **6**, 39 (2012)
33. Schirrmeister, R.T., Springenberg, J.T., Fiederer, L.D.J., Glasstetter, M., Eggensperger, K., Tangermann, M., Hutter, F., Burgard, W., Ball, T.: Deep learning with convolutional neural networks for EEG decoding and visualization. Human Brain Mapping **38**(11), 5391–5420 (2017)